\documentclass[conference]{IEEEtran}
\IEEEoverridecommandlockouts
\usepackage{cite}
\usepackage{amsmath,amssymb,amsfonts}
\usepackage{algorithmic}
\usepackage{graphicx}
\usepackage{textcomp}
\usepackage{subfig}
\usepackage[dvipsnames, table]{xcolor}
\usepackage{soul}
\usepackage{tikz}
\usetikzlibrary{circuits.logic.US}
\usetikzlibrary{shapes.geometric}
\usetikzlibrary{decorations.pathreplacing}
\usepackage{multicol}
\tikzset{every picture/.style={/utils/exec={\sffamily}}}
\tikzset{
  multiplexer/.style={
    draw,
    trapezium,
    shape border uses incircle,
    shape border rotate=270,
    minimum size=14pt,
  }
}
\tikzstyle{branch}=[fill,shape=circle,minimum size=3pt,inner sep=0pt]
\newcommand{\code}[1]{\texttt{#1}}
\usepackage{booktabs}
\usepackage{multirow}
\usepackage{orcidlink}
\definecolor{DarkGreen}{RGB}{0,80,0}
\definecolor{AppleGreen}{rgb}{0.55, 0.71, 0.0}
\usepackage{amsmath, amssymb}

\def\BibTeX{{\rm B\kern-.05em{\sc i\kern-.025em b}\kern-.08em
    T\kern-.1667em\lower.7ex\hbox{E}\kern-.125emX}}

\newcommand{\filterblind}[1]{%
#1
}

\begin{document}

\title{
Shedding the Bits: Pushing the Boundaries of Quantization with Minifloats on FPGAs
\thanks{Work by Shivam Aggarwal and Hans Jakob Damsgaard was carried out during internships with AMD Research.}
}

\author{
    \IEEEauthorblockN{Shivam Aggarwal\IEEEauthorrefmark{1}\IEEEauthorrefmark{2}\orcidlink{0000-0003-1748-9810}, Hans Jakob Damsgaard\IEEEauthorrefmark{3}\IEEEauthorrefmark{2}\orcidlink{0000-0001-8409-0282}, Alessandro Pappalardo\IEEEauthorrefmark{2}\orcidlink{0000-0001-9386-5510}, Giuseppe Franco\IEEEauthorrefmark{4},\\Thomas B. Preu{\ss}er\IEEEauthorrefmark{2}\orcidlink{0000-0003-3998-7896}, Michaela Blott\IEEEauthorrefmark{2}\orcidlink{0000-0002-7833-4057}, Tulika Mitra\IEEEauthorrefmark{1}\orcidlink{0000-0003-4136-4188}}
    \IEEEauthorblockA{\IEEEauthorrefmark{1}School of Computing, National University of Singapore, Singapore, \IEEEauthorrefmark{2}AMD Research and \\ Advanced Development, Dublin, Ireland, \IEEEauthorrefmark{3}Electrical Engineering Unit, Tampere University, Finland, \IEEEauthorrefmark{4}AMD, Germany}
}

\maketitle

\begin{abstract}
Post-training quantization (PTQ) is a powerful technique for model compression, reducing the numerical precision in neural networks without additional training overhead. Recent works have investigated adopting 8-bit floating-point formats (\code{FP8}) in the context of PTQ for model inference. However, 
floating-point formats smaller than 8 bits and their relative comparison in terms of accuracy-hardware cost with integers remains unexplored on FPGAs. 
In this work, we present minifloats, which are reduced-precision floating-point formats capable of further reducing the memory footprint, latency, and energy cost of a model while approaching full-precision model accuracy. 
We implement a custom FPGA-based multiply-accumulate operator library and explore the vast design space, comparing minifloat and integer representations across 3 to 8 bits for both weights and activations. We also examine the applicability of various integer-based quantization techniques to minifloats. Our experiments show that minifloats offer a promising alternative for emerging workloads such as vision transformers.
\end{abstract}


\section{Introduction}
With the increasing demand for deploying machine learning models on resource-constrained devices, post-training quantization (PTQ)~\cite{Nagel2019DataFreeQT, adaround, frantar-gptq} has been the prevailing choice for model compression, offering high model accuracy with reduced compute and memory usage. PTQ enables 
low-precision arithmetic without any model re-training and with minimal fine-tuning or calibration using only a small unlabeled dataset. 

Multiply-accumulate (MAC) operations are ubiquitous in deep learning models. While GPUs with their high concentration of MACs remain the primary computing platform for these workloads, their support for different numerical precisions is inherently limited. In contrast, FPGAs offer unparalleled flexibility by supporting arbitrary formats at the bit level, presenting a unique advantage. Lately, several quantization works~\cite{fp8, Kuzmin2022FP8QT, noune20228bit} have explored the idea of utilizing 8-bit floating-point (\code{FP8}) representations in lieu of integers. Integer formats are often preferred over floating-point alternatives due to their simpler hardware implementation. However, as the bit-width of floating-point formats decreases, so do their resource footprints. This leads us to hypothesize floating-point MACs with size and throughput akin to their integer counterparts.

For instance, prior research by Xilinx demonstrated a 7-bit floating-point representation that reduces resource consumption significantly while maintaining comparable or even superior accuracy in comparison to the \code{INT8} format~\cite{Xilinx}. This observation makes reduced-precision floating-point arithmetic on FPGAs a practical and valuable choice, especially for models like transformers~\cite{dai-etal-2019-transformer}, which frequently feature numerous outliers in their activation distribution~\cite{xiao2023smoothquant}.

Building upon these advancements, our study delves into the realm of \textit{minifloat quantization} and explores the hardware efficiency of representations with fewer than 8~bits on FPGAs. Such minifloats with low bit-widths can exhibit better model performance compared to their integer counterparts, mostly due to their superior dynamic range with more values close to zero. Moreover, the flexibility to adjust both the number of exponent and mantissa bits allows us to mitigate the loss in model accuracy that often accompanies reduced precision.

Unfortunately, existing literature predominantly focuses on reduced-precision integer quantization, with limited exploration of minifloats in terms of accuracy-hardware tradeoffs. To address this gap, we undertake a comprehensive investigation of the design space encompassing minifloats and integer quantization. Our investigation spans precisions ranging from 3 to 8~bits, applied to both weights and activations across a set of deep learning vision workloads for the ImageNet classification task~\cite{deng2009imagenet}. 

Furthermore, we implement custom bit-width MAC units for both minifloats and integers to analyze the impact of the number of exponent and mantissa bits within different minifloat formats on FPGAs. Finally, we investigate the effect of existing post-training optimization techniques, such as SmoothQuant~\cite{xiao2023smoothquant}, gradient-based learned rounding~\cite{adaround}, and GPTQ~\cite{frantar-gptq} for minifloats. Our main contributions are:

\begin{itemize}
    \item We propose a novel PTQ quantization framework for low-precision minifloats, ranging from 3 to 8 bits. 
    \item We implement an operator library to realize custom bit-width integer and minifloat MACs on FPGAs.
    \item We thoroughly explore the accuracy-hardware trade-offs, providing an in-depth analysis of three prominent vision models -- ResNet-18, MobileNetV2, and ViT-B-32 -- based on our custom FPGA-based operator library.
\end{itemize}

Our experiments indicate that minifloat quantization typically outperforms integer quantization for bit-widths of four or more, both for weights and activations. However, when compared against our FPGA hardware cost model, integer quantization often retains its Pareto optimality due to its slightly smaller hardware footprint than minifloats at a given~precision.

\section{Related Work}
Recent research proposes the adoption of \code{FP8} formats for efficient model inference. 
Micikevicius~\textit{et~al.}~\cite{fp8} introduce two \code{FP8} formats
to represent the weights and activations of a neural network. 
Kuzmin~\textit{et~al.}~\cite{Kuzmin2022FP8QT} and Noune~\textit{et~al.}~\cite{noune20228bit} study the effects of the \code{FP8} format, varying the number of exponent and mantissa bits, and the exponent bias. They observe significant performance improvements by searching for the best exponent bias term instead of setting the bias 
as per the IEEE 754 standard~\cite{4610935}. More complex group-wise quantization techniques~\cite{dettmers2023case} have also been proposed for floating-point formats smaller than 8~bits, targeting specifically large language models (LLMs). However, these methods maintain actual computation in the full-precision format without considering any hardware constraints. 

Numerous studies~\cite{10.1145/3546182, 5377624, Vstias2019HybridDC} explore optimized MAC designs for deep learning on FPGAs. Most of these works focus on 8-bit fixed-point quantization~\cite{Vstias2019HybridDC, 10.1145/3373087.3375311}, with a few considering binary quantization that converts MACs into popcounts~\cite{8532584, finn}. A few works propose accurate floating-point MACs for \code{FP32} or \code{FP64} operands~\cite{5377624, wilson2016unified, kadric2016accurate}. Other studies investigate quantization for low-precision accumulation~\cite{a2q, 8966322}. Finally, FloPoCo offers a flexible library for a wide range of arithmetic units for FPGAs, including floating-point MACs, but limit designs to have heterogeneous operand formats~\cite{de2008fpga}. In contrast to existing works, we systematically analyze various configuration settings for reduced-precision integer and minifloat quantization on FPGAs using custom bit-width MACs. Our study provides a detailed perspective on the considerations involved in selecting the appropriate precision formats with the target neural network, model accuracy, and various hardware constraints in mind.


\section{Background: Integer Quantization} 
\begin{figure*}%
    \centering
    \subfloat[\centering \code{INT4} Quantization.\label{fig:intquant}]{{\includegraphics[width=0.50\linewidth]{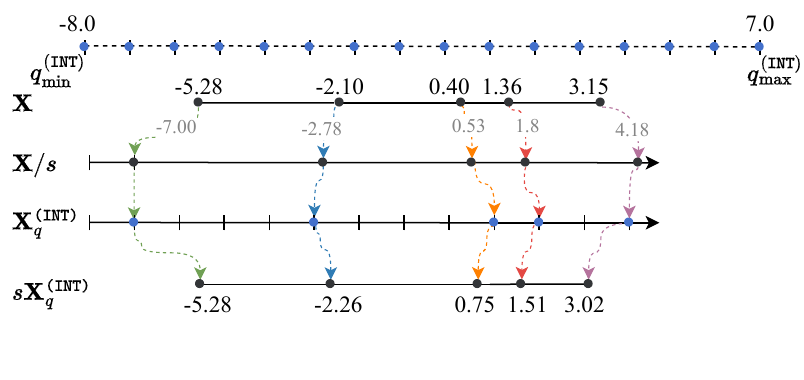} }}%
    \hfill
    \subfloat[\centering \code{E2M1} Quantization.\label{fig:fpquant}]{{\includegraphics[width=0.44\linewidth]{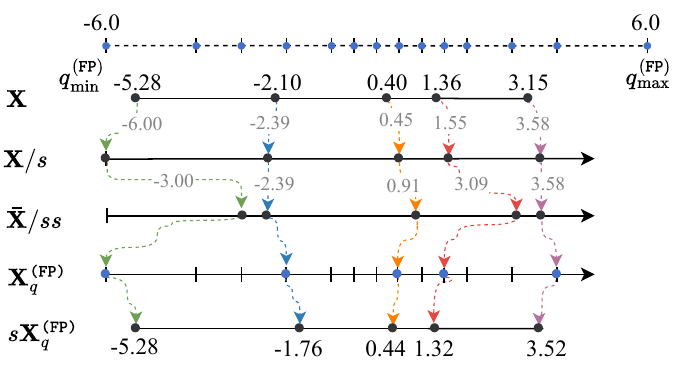} }}%
    \caption{PTQ process with \code{INT4} representation and \code{E2M1} representation.}%
    \label{fig:quant}%
       \vskip -0.23in
\end{figure*}

Table~\ref{tab:notation} outlines the notations used in this work. For integer (\code{INT}) quantization, given a tensor $\mathbf{X}$, associated scaling factor $\mathtt{s}$, and zero-point $\mathtt{z}$, the quantization Eq.~(\ref{eq1_int}) and dequantization Eq.~(\ref{eq2_int}) operations are defined as follows: 

\noindent
\begin{equation}
\resizebox{0.91\hsize}{!}{$
\mathbf{X}_q =     \operatorname{quantize}(\mathbf{X}; \mathtt{s}, \mathtt{z}) 
           = \operatorname{clip} \left( \bigg 
\lfloor \frac{\mathbf{X}}{\mathtt{s}} \bigg \rceil + \mathtt{z};\; 
q_\text{min}^{(\code{INT})},  q_\text{max}^{(\code{INT})} 
\right) 
$}
\label{eq1_int}
\end{equation}
\noindent
\begin{equation}
    \operatorname{dequantize}(\mathbf{X}_q; \mathtt{s}, \mathtt{z}) = \mathtt{s} \cdot (\mathbf{X}_q - \mathtt{z})
\label{eq2_int}
\end{equation}

\noindent where $\lfloor\cdot\rceil$ is the round-to-nearest operator and $\mathtt{s}$ is defined as: 
\noindent
\begin{equation}
 \mathtt{s} = \frac{{t}}{q_\text{max}^{(\code{INT})}} 
 \label{eq_int_scale}
\end{equation}

\noindent over the entire tensor \textbf{(per-tensor)} or for each output channel in the tensor \textbf{(per-channel)}. For weights, we define $t = \operatorname{max}(\lvert \mathbf{X} \rvert)$ or ${t}_{j}$ = $\operatorname{max}(\lvert \mathbf{X}_{j}\rvert)$, $j$ = 1, 2, $\dots$, $C_o$, where $j$ refers to the $j$-th channel and $C_o$ corresponds to the total number of output channels. For activations, $t$ is a per-tensor value determined through a calibration procedure.
In general, $[q_\text{min}^{(\code{INT})}, q_\text{max}^{(\code{INT})}]$ is the quantization range controlled by the target bit-width $r$. For signed integers, $q_\text{min}^{(\code{INT})}$ = $-2^{r - 1}$ and $q_\text{max}^{(\code{INT})}$ = $2^{r - 1} - 1$. For unsigned integers,  $q_\text{min}^{(\code{INT})}$ = $0$ and $q_\text{max}^{(\code{INT})}$ = $2^{r} - 1$. For both weights and activations, we keep $z = 0$. 
Fig.~\ref{fig:intquant} describes the complete integer quantization flow for the \code{INT4} format.

\begin{table}[t]
\centering
\caption{Notation used in this work.}
\label{tab:notation}
\begin{tabular}{@{}lc@{}}
\toprule
\textbf{Name}                  & \textbf{Notation}                                            \\ \hline
tensors (general, weights, activations)                         & $\mathbf{X}$, ${\mathbf{W}}$, ${\mathbf{Y}}$                                                 \\ \hline
tensor value                  & $\mathbf{{x}}$                                           \\ \hline
scaled tensor value                  & $\mathbf{\overline{x}}$                                           \\ \hline
quantized tensor                         & $\mathbf{X_q}$                                                 \\ \hline
scaling factor                 & $\mathtt{s}$                                                          \\ \hline
zero-point                     & $\mathtt{z}$                                                          \\ \hline
maximum value of tensor              & $t$                                                          \\ \hline
quantization range                      & $[q_\text{min}, q_\text{max}]$ \\ \hline
\code{FP} format fields (sign, exponent, mantissa)                       & $S$, $E$, $M$                                                          \\ \hline
\code{FP} mantissa bit-width                  & $m$                                                          \\ \hline
\code{FP} exponent bit-width                  & $e$                                                          \\ \hline
total bit-width                  & $r$                                                          \\ \hline
exponent bias                  & $b$                                                          \\ \hline
exponent value       & $u$                                                          \\ \hline
fine-grained, internal scale                 & $\mathtt{ss}$                                                         \\ \bottomrule
\end{tabular}
\vskip -0.25in
\end{table}

\section{Minifloat Quantization}
A standard IEEE floating-point (\code{FP}) format consists of three key components: a sign bit $S \in \{0, 1\}$, an $m$-bit mantissa $M$, and an $e$-bit exponent $E$~\cite{4610935}. These elements sum up to the total bit-width of the format $r$. Given the exponent bias $b$ as the final parameter of the format, a normalized floating-point representation is interpreted by:
\noindent
\begin{equation}
 {x}^{(\code{FP})} = (-1)^{S} \times 2^{u - b} \times \left(1 + \underset{i = 1}{\sum^{m}} {M_{i}} \times {2^{-i}}\right) 
 \label{fp_format}
\end{equation}

\noindent where $0 \leq u < 2^{e}$ represents the range of exponent values and $M_i \in \{0, 1\}$ denotes the $i$\textsuperscript{th} bit of the $m$-bit mantissa. The mantissa combined with the implicit digit of the floating-point number constitutes its \textit{significand}. The IEEE standard \code{FP32} format~\cite{4610935} features $e = 8$ and $m = 23$, with an exponent bias defined as $b = 2^{e-1} - 1$, which equals 127. The number of mantissa bits determines the precision of values within a given range, the number of exponent bits governs the dynamic range of representable values, and the exponent bias controls the position of the range on the real number line. In general, $[q_\text{min}^{(\code{FP})}, q_\text{max}^{(\code{FP})}]$ is the quantization range determined by the number of exponent and mantissa bits and the exponent bias. Here, $q_\text{min}^{(\code{FP})}$ $=$ $-(2 - 2^{-m}) \cdot 2^{2^e - b - 1}$ and $q_\text{max}^{(\code{FP})}$ $=$ $(2 - 2^{-m}) \cdot 2^{2^e - b - 1}$.

In this study, we investigate \textit{minifloats}, which are reduced-precision floating-point representations ranging from 3 to 8 bits. Later, we denote \code{FP} formats with $e=x$ and $m=y$ by \code{ExMy}. We adhere to IEEE standards for the exponent bias and set it to $b = 2^{e-1} - 1$. However, we deviate from IEEE conventions on the treatment of \code{inf} and \code{NaN} and represent neither. We choose not to represent \code{inf} as we assume that the values outside the representation range of a minifloat are saturated. Once \code{inf} is excluded, \code{NaN} cannot be generated when performing only multiplications and additions. Our design space exploration for various floating-point representations is guided by the constraints: $e \in [{1, r - 1})$ and $m = r - 1 - e$. We do take advantage of subnormal numbers to maintain precision for values near 0. Subnormal numbers have an exponent value $u$ set to 1 and are interpreted to have a leading significand digit of zero rather than the one shown in Eq.~(\ref{fp_format}). This allows for extra dynamic range and a graceful degradation of significand precision by introducing leading zeros.

We present a novel minifloat quantization method with two-level scaling, as illustrated in Fig.~\ref{fig:fpquant}. Several recent works emphasize the careful selection of appropriate scaling factors and exponent biases for \code{FP8} formats~\cite{Kuzmin2022FP8QT, perez2023training}. We take a more straightforward approach that is aligned with existing integer quantization techniques. We introduce a coarse-grained scaling factor $s$ maintained \textbf{per-tensor} or \textbf{per-channel} to normalize the original tensor. Formally, for tensor $\mathbf{X}$ and scaling factor~$\mathtt{s}$:
\begin{multicols}{2}
\noindent
\begin{equation}
 {{\mathtt{s}}} = \frac{{t}}{q_\text{max}^{(\code{FP})}} 
\label{eq1_fp}
\end{equation}
\noindent
\begin{equation}
 {\mathbf{\overline{X}}} = \frac{\mathbf{X}}{{{\mathtt{s}}}} 
\label{eq2_fp}
\end{equation}
\end{multicols}
In accordance with the integer quantization process, we also define for weights $t = \operatorname{max}(\lvert \mathbf{X} \rvert)$ or ${t}_{j}$ = $\operatorname{max}(\lvert \mathbf{X}_{j}\rvert)$, $j$ = 1, 2, $\dots$, $C_o$, where $j$ refers to the $j$-th channel and $C_o$ corresponds to the total number of output channels. For activations, $t$ is a per-tensor value determined through the calibration procedure.

We maintain a fine-grained, internal scaling factor $\mathtt{ss}$ for each element $\mathbf{\overline{x}}$ in the scaled tensor $\mathbf{\overline{X}}$ as in~\cite{Kuzmin2022FP8QT}. This scaling factor depends on the number of mantissa bits $m$ and the specific range the element $\mathbf{\overline{x}}$ falls into. The scaling factor $\mathtt{ss}$ maps each element $\mathbf{\overline{x}}$ onto its nearest neighbor within the minifloat quantization grid. It is maintained as a power-of-two and can be stored with minimal overhead. The expression $2^{1 - b - m}$ serves as the threshold for the smallest value representable by a given minifloat format. Consequently, any value $p$ that falls below this threshold is subjected to a clipping operation and subsequently assigned a value of $1 - b - m$. Given the specified values of $\mathbf{\overline{x}}$, $m$, and $b$, we define this condition as follows: 
\noindent
\begin{equation}
\log_2 {\mathtt{ss}} = {p} = \operatorname{max}(\lfloor\log_2|\mathbf{\overline{x}}|\rfloor - m, 1 - b - m)
\label{eq3_fp}
\end{equation}

\noindent Finally, we define the minifloat quantization operation Eq.~(\ref{eq4_fp}) for $\mathbf{\overline{x}}$ and its associated dequantization operation Eq.~(\ref{eq5_fp})~as: 
\noindent
\begin{equation}
\begin{split}
     \mathbf{x}_q &= \operatorname{quantize}(\mathbf{\overline{x}}; e, m, b, \mathtt{s}, \mathtt{ss})  \\
           &= \operatorname{clip} \left({\mathtt{ss}} \bigg 
\lfloor \frac{\mathbf{\overline{x}}}{{\mathtt{ss}}} \bigg \rceil;\; q_\text{min}^{(\code{FP})},  q_\text{max}^{(\code{FP})}\right) 
\label{eq4_fp}
\end{split}
\end{equation}
\noindent
\begin{equation}
    \operatorname{dequantize}(\mathbf{x}_q; \mathtt{s}) = \mathtt{s} \cdot \mathbf{x}_q
\label{eq5_fp}
\end{equation}

\noindent where $\lfloor\cdot\rceil$ is the round-to-nearest operator.

\subsection{PTQ Optimization for Minifloats}
In this section, we discuss several \textit{integer} PTQ methods and their applicability to minifloat quantization. 


\paragraph{SmoothQuant}
Activation quantization can be challenging especially because of outliers, which are difficult to capture with static per-tensor scaling factors. SmoothQuant~\cite{xiao2023smoothquant} addresses this challenge by introducing per-channel smoothing factors, thus distributing the burden of quantization between input activations and the following weight matrix. This integration, facilitated by scaling before quantization, seamlessly aligns with the proposed minifloat quantization flow.

\paragraph{Bias Correction} Another major challenge in quantization is the introduction of biased errors within the output distribution of a layer. These biases primarily stem from the errors introduced by both weights and activations in the preceding layer,
ultimately leading performance degradation, as highlighted in~\cite{Nagel2019DataFreeQT, pmlr-v97-meller19a, Finkelstein2019FightingQB}. In this work, we study the impact of \textit{empirical bias correction} for minifloat quantization by analyzing the layer-wise quantization errors using a small calibration dataset. By subtracting the product of these errors and the mean value of the input activations from the biased output, we aim to alleviate the detrimental effects of biased errors within subsequent layers.


\paragraph{Gradient-based Learned Rounding} \label{sec:lr} The integer and minifloat quantization processes rely on the round-to-nearest operator $\lfloor\cdot\rceil$, as shown in Eq.~(\ref{eq1_int}) and Eq.~(\ref{eq4_fp}), to project full-precision \code{FP32} values onto their nearest neighbor in the quantization grid. However, previous works~\cite{adaround, Li2021BRECQPT} underscore the sub-optimality of this operation in the context of post-training \textit{integer} quantization.
Building upon these insights, we observe a similar scenario with the minifloat quantization, where the round-to-nearest operation may be sub-optimal under certain circumstances. To address this, we advocate adopting a gradient-based learned rounding approach akin to the principles expounded in the literature. Specifically, we adapt Eq.~(\ref{eq4_fp}) for the scaled weight values $\mathbf{\overline{w}}$ as follows:
\noindent
\begin{equation}
\begin{split}
     {\mathbf{{w}}_q^{(\code{FP})}} &= \operatorname{quantize}(\mathbf{\overline{w}}; e, m, b, \mathtt{s}, \mathtt{ss})  \\
           &= \operatorname{clip} \left(\mathtt{{ss}} \bigg 
\lfloor \frac{\mathbf{\overline{w}}}{\mathtt{{ss}}} \bigg \rfloor + h(\mathbf{V}) ;\; q_\text{min}^{(\code{FP})},  q_\text{max}^{(\code{FP})}\right) 
\end{split}
\label{eq5_lr}
\end{equation}

\noindent Here, $h(\mathbf{V})$ is a rectified sigmoid function~\cite{DBLP:conf/iclr/LouizosWK18} using the learnable variable $\mathbf{V}$, and hyperparameters $\zeta$ $(=$$1.1)$ and $\gamma$ $(=$$-0.1)$. To minimize the overall objective, the regularization function encourages the term $h(\mathbf{V})$ to converge on 0 or 1, forcing the round-to-nearest operator to approximate to the floor or ceil function. All other hyperparameters remain consistent with those employed in integer-based learned rounding.

\paragraph{GPTQ} \label{sec:gptq} 
Lastly, we investigate applying GPTQ~\cite{frantar-gptq} to minifloat quantization. GPTQ leverages tricks such as unordered quantization, local updates, and Cholesky decomposition to efficiently quantize large-scale models at very low precision. Remarkably, the minifloat quantization formulation seamlessly integrates within the existing GPTQ framework, requiring no additional hyperparameter fine-tuning.


\section{FPGA Operator Library}

\begin{figure}[t]
    \centering
    \subfloat[Integer MAC.\label{sfig:imacc}]{\begin{tikzpicture}[circuit logic US, scale=.8]
\coordinate (BASE) at (0,0);
\node [circle,draw,thick,minimum width=18pt,scale=.8,fill=BrickRed!40] (mult) at (BASE) {$\pmb{\times}$};
\node (a) at ($(mult.west)-(.35,-.5)$) {$\textbf{a}$};
\node (b) at ($(mult.west)-(.35,.5)$)  {$\textbf{b}$};
\node [circle,draw,thick,minimum width=18pt,scale=.8,fill=BrickRed!40] (add) at ($(mult.east)+(.6,0)$) {$\textbf{+}$};
\node [multiplexer,thick,rotate=-90,scale=.8,fill=gray!40] (mux) at ($(add.north)+(0,.5)$) {};
\node (0) at ($(mux.south west)+(0,.5)$) {$\textbf{0}$};
\node [rectangle,draw,thick,minimum width=16pt,minimum height=32pt,fill=BurntOrange!40] (reg) at ($(add.east)+(.75,0)$) {Acc};
\node (p) at ($(reg.east)+(.75,0)$) {};
\draw ($(reg.south)-(4pt,0)$) -- ($(reg.south)+(0,8pt)$) -- ($(reg.south)+(4pt,0)$); 

\draw (a) -- (mult);
\draw (b) -- (mult);
\draw (mult) -- (add);
\draw (add) -- (reg.west);
\draw (reg.east) -- (p);
\draw ($(reg.east)+(.25,0)$) |- ($(mux.north west)+(0,.25)$) -- (mux.north west);
\draw (mux.east) -- (add);
\draw (0) -- (mux.south west);
\end{tikzpicture}}
    \hfill
    \subfloat[Minifloat MAC.\label{sfig:fmacc}]{\begin{tikzpicture}[circuit logic US, scale=.8]
\coordinate (BASE) at (0,0);

\node [circle,draw,thick,minimum width=18pt,inner sep=2pt,outer sep=0pt,scale=.8,fill=CadetBlue!50] (compa) at ($(BASE)-(-1,-1)$) {$\pmb{\neq\!0}$};
\node (a) at ($(compa.west)-(1.6,0)$) {};
\node (sa) at ($(a)+(.5,.35)$) {$\textbf{S}_a$};
\node (ea) at ($(a)+(.5,0)$) {$\textbf{E}_a$};
\node (ma) at ($(a)+(.5,-.35)$) {$\textbf{M}_a$};
\node [circle,draw,thick,minimum width=18pt,inner sep=2pt,outer sep=0pt,scale=.8,fill=CadetBlue!50] (compb) at ($(BASE)-(-1,1)$) {$\pmb{\neq\!0}$};
\node (b) at ($(compb.west)-(1.6,0)$) {};
\node (sb) at ($(b)+(.5,.35)$) {$\textbf{S}_b$};
\node (eb) at ($(b)+(.5,0)$) {$\textbf{E}_b$};
\node (mb) at ($(b)+(.5,-.4)$) {$\textbf{M}_b$};
\draw (ea) -- ++(.55,0); \draw ($(ea)+(.7,0)$) -- (compa);
\draw (eb) -- (compb);

\node [circle,draw,thick,minimum width=18pt,scale=.8,fill=BrickRed!40] (mult) at ($(compb.east)+(1.15,0)$) {$\pmb{\times}$};
\node [circle,draw,thick,minimum width=18pt,scale=.8,fill=BrickRed!40] (shift) at ($(mult.east)+(.65,0)$) {$\pmb{\ll}$};
\draw (ma) -- ++(.55,-.55); \draw ($(ma)+(.7,-.7)$) -- ++(.1,-.1); \draw ($(ma)+(1,-1)$) -- ++(.1,-.1) -- ++(.8,0); \draw ($(ma)+(2.15,-1.1)$) -- ++(.2,0) node [inner sep=0pt,outer sep=0pt] (norma) {} -| ($(mult.west)+(-.25,.25)$) -- (mult);
\draw ($(norma)$) -- ++(-.1,.1) -- ++(0,.1) node [inner sep=0pt,outer sep=0pt] (normaoffset) {}; \draw ($(normaoffset)+(0,.2)$) -- ++(0,.3); \draw ($(normaoffset)+(0,.7)$) -- ++(0,.55);
\draw (mb) -- ++(.5,-.2) |- ($(compb.east)+(.25,-.6)$) node [inner sep=0pt,outer sep=0pt] (normb) {} -| ($(mult.west)-(.25,.25)$) -- (mult);
\draw (normb) -- ++(-.1,.1) -- ++(0,.5);
\draw (mult) -- (shift);

\node [circle,draw,thick,minimum width=18pt,scale=.8,fill=BrickRed!40] (expadd) at ($(shift)+(0,1.7)$) {$\textbf{+}$};
\node [circle,draw,thick,minimum width=18pt,scale=.8,fill=BrickRed!40] (expasub) at ($(mult)+(0,2.25)$) {$\pmb{-}$};
\node [circle,draw,thick,minimum width=18pt,scale=.8,fill=BrickRed!40] (expbsub) at ($(mult)+(0,1.1)$) {$\pmb{-}$};
\draw (expasub) -- (expadd);
\draw (expbsub) -- (expadd);
\draw (expadd) -- (shift);
\draw ($(ea)+(.9,0)$) |- ($(expasub.west)+(-.25,.25)$) -- (expasub);
\draw (compa.east) -- ($(expasub.west)-(.25,.25)$) -- (expasub);
\draw ($(eb)+(.9,0)$) |- ($(expbsub.west)+(-.25,.25)$) -- (expbsub);
\draw (compb.east) -- ++(.15,0) |- ($(expbsub.west)-(.25,.25)$) -- (expbsub);

\node [multiplexer,thick,scale=.8,fill=gray!40] (invmux) at ($(shift.east)+(1.25,0)$) {};
\node [circle,draw,thick,minimum width=5pt,inner sep=0pt,outer sep=0pt] (invball) at ($(invmux.south west)-(.25,0)$) {};
\coordinate (invin) at ($(invball.west)-(8pt,0)$);
\draw [thick] (invin) -- ($(invin)+(0,5pt)$) -- (invball.west) -- ($(invin)-(0,5pt)$) -- (invin);
\draw (shift.east) -- ($(shift.east)+(.2,0)$) |- (invmux.north west);
\draw ($(shift.east)+(.2,0)$) |- (invin);
\draw (invball.east) -- (invmux.south west);

\node [xor gate,thick,rotate=-90,scale=.75] (xor) at ($(invmux)+(0,2)$) {};
\draw (xor.east) -- (invmux.north);
\draw (sa) -- ++(.4,.4) |- ([xshift=-3pt]$(xor.north west)+(0,.55)$) -- ([xshift=-3pt,yshift=-1.5pt]xor.north west);
\draw (sb) -- ++(.65,.65) |- ([xshift=3pt]$(xor.south west)+(0,.35)$) -- ([xshift=3pt,yshift=-1.5pt]xor.south west);

\node [circle,draw,thick,minimum width=18pt,scale=.8,fill=BrickRed!40] (add) at ($(invmux.east)+(.6,0)$) {$\textbf{+}$};
\coordinate (diff) at ($(xor.east)-(invmux.north)$);
\draw ($(xor.east)-0.6*(diff)$) -- (add);
\node [multiplexer,thick,rotate=-90,scale=.8,fill=gray!40] (mux) at ($(add.north)+(0,.5)$) {};
\node (0) at ($(mux.south west)+(0,.5)$) {$\textbf{0}$};
\node [rectangle,draw,thick,minimum width=16pt,minimum height=32pt,fill=BurntOrange!40] (reg) at ($(add.east)+(.75,0)$) {Acc};
\node (p) at ($(reg.east)+(.75,0)$) {$\textbf{}$};
\draw (invmux.east) -- (add.west);
\draw ($(reg.south)-(4pt,0)$) -- ($(reg.south)+(0,8pt)$) -- ($(reg.south)+(4pt,0)$); 
\draw (add) -- (reg.west);
\draw (reg.east) -- (p);
\draw ($(reg.east)+(.25,0)$) |- ($(mux.north west)+(0,.25)$) -- (mux.north west);
\draw (mux.east) -- (add);
\draw (0) -- (mux.south west);

\end{tikzpicture}}
    \caption{Simplified illustrations of the considered integer and minifloat MACs with two operands, $\textbf{a}$ and $\textbf{b}$.}
    \label{fig:maccs}
       \vskip -0.27in
\end{figure}
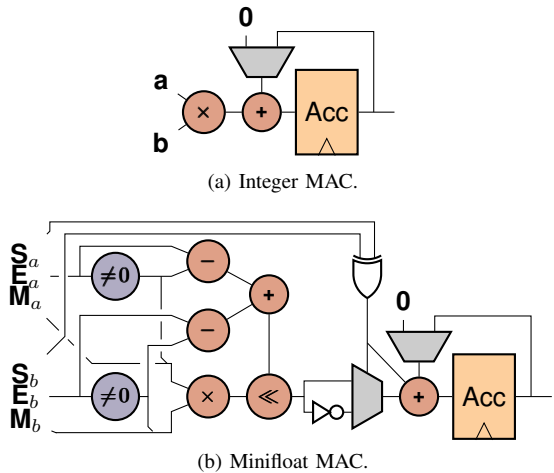

To explore our hypothesis on minifloat hardware efficiency, we implement MACs for integer and minifloat representations on a recent AMD Versal\filterblind{™} VCK190 device. Our integer MAC shown in Fig.~\ref{sfig:imacc} is trivial, whereas our minifloat MAC is more complex and involves a long fixed-point accumulator~\cite{kulisch2011exact}. As Fig.~\ref{sfig:fmacc} shows, the two minifloat operands undergo several preliminary operations~--~primarily, the extraction and subsequent multiplication and shifting of their significands~--~before they can be added to the accumulator. The sign-inversion of the mantissa products, shown with an inverter and a multiplexer, is merged with the subsequent addition. This architecture is modeled after the well-established \textit{long adder and shift} design from~\cite[Section~8.4.1]{kulisch2008computer} and~\cite[Section~2.2]{de2017design}. Unlike recent work that applies segmentation to the accumulator for improved latency~\cite[Sections~8.4.2-4]{kulisch2008computer}~\cite{de2017design}, we maintain the long adder as it maps well to the hardened carry primitives available in the AMD FPGAs~\cite{xilinx2023ds950}. 

\begin{table*}[t]
\centering
\caption{Top-1 model accuracy (\%) and hardware utilization (\#LUTs) results for different floating-point (\code{FP}) and integer (\code{INT}) representations compared to the full-precision \code{FP32} model in ascending order of the weights and activation bit-widths. Weight and activation bit-widths are denoted by \code{W}$\times$\code{A}. The best model accuracies are highlighted in \textbf{\colorbox{DarkGreen!25}{{\textcolor{DarkGreen}{Green}}}}, while those closely trailing the best model, with a difference of less than $0.05$, are highlighted in {\colorbox{AppleGreen!25}{{Olive Green}}}.}
\label{fp_int_acc}
\setlength\tabcolsep{5pt}
\begin{tabular}{@{}c||c||ccccc||ccccc||ccccc@{}}
\toprule
\multirow{3}{*}{\begin{tabular}[c]{@{}c@{}}\\\#config \\ \code{W}$\times$\code{A}\end{tabular}} &  & \multicolumn{5}{c||}{\textbf{ResNet-18}} & \multicolumn{5}{c||}{\textbf{MobileNetV2}} & \multicolumn{5}{c}{\textbf{ViT-B-32}} \\ \cmidrule(l){3-17}  
 & \#LUTs & \multicolumn{2}{c|}{{Acc (\%)}} & \multicolumn{3}{c||}{\code{FP} config} & \multicolumn{2}{c|}{{Acc (\%)}} & \multicolumn{3}{c||}{\code{FP} config} & \multicolumn{2}{c|}{{Acc (\%)}} & \multicolumn{3}{c}{\code{FP} config} \\ \cmidrule(l){3-17}  
 & \code{INT} & \code{INT} & \multicolumn{1}{c|}{\code{FP}} & W & A & \#LUTs & \code{INT} & \multicolumn{1}{c|}{\code{FP}} & W & A & \#LUTs & \code{INT} & \multicolumn{1}{c|}{\code{FP}} & W & A & \#LUTs \\ \midrule
 
\code{FP32} & $-$ & \multicolumn{2}{c|}{69.76} & \multicolumn{2}{c}{\code{E8M23}} & $-$ & \multicolumn{2}{c|}{71.90} & \multicolumn{2}{c}{\code{E8M23}} & $-$ & \multicolumn{2}{c|}{75.91} & \multicolumn{2}{c}{\code{E8M23}} & $-$ \\ \midrule

\code{3$\times$3} & \phantom{0}25 & \cellcolor{DarkGreen!25}{\textcolor{DarkGreen}{\textbf{48.22}}} & \multicolumn{1}{c|}{46.0\phantom{0}} & \code{E1M1} & \code{E1M1} & \phantom{0}27 & {\phantom{0}0.21} & \multicolumn{1}{c|}{\phantom{0}0.17} & \code{E1M1} & \code{E1M1} & \phantom{0}27 & \phantom{0}0.11 & \multicolumn{1}{c|}{\phantom{0}0.11} & \code{E1M1} & \code{E1M1} & \phantom{0}27 \\
\code{3$\times$4} & \phantom{0}37 & 62.36 & \multicolumn{1}{c|}{\cellcolor{DarkGreen!25}{\textcolor{DarkGreen}{\textbf{62.50}}}} & \code{E1M1} & \code{E2M1} & \phantom{0}42 & {\phantom{0}5.13} & \multicolumn{1}{c|}{\phantom{0}3.93} & \code{E1M1} & \code{E2M1} & \phantom{0}42 & \cellcolor{DarkGreen!25}{\textcolor{DarkGreen}{\textbf{33.45}}} & \multicolumn{1}{c|}{\phantom{0}0.11} & \code{E1M1} & \code{E1M1} & \phantom{0}27 \\
\code{3$\times$5} & \phantom{0}38 & {\cellcolor{DarkGreen!25}{\textcolor{DarkGreen}{\textbf{65.95}}}} & \multicolumn{1}{c|}{65.48} & \code{E1M1} & \code{E2M2} & \phantom{0}44 & 25.38 & \multicolumn{1}{c|}{\cellcolor{DarkGreen!25}{\textcolor{DarkGreen}{\textbf{25.57}}}} & \code{E1M1} & \code{E2M2} & \phantom{0}44 & 67.16 & \multicolumn{1}{c|} {\cellcolor{DarkGreen!25}{\textcolor{DarkGreen}{\textbf{69.29}}}} & \code{E1M1} & \code{E3M1} & \phantom{0}54 \\
\code{4$\times$4} & \phantom{0}40 & 64.71 & \multicolumn{1}{c|}{\cellcolor{DarkGreen!25}{\textcolor{DarkGreen}{\textbf{66.48}}}} & \code{E2M1} & \code{E2M1} & \phantom{0}43 & 17.50 & \multicolumn{1}{c|}{\cellcolor{DarkGreen!25}{\textcolor{DarkGreen}{\textbf{21.83}}}} & \code{E1M2} & \code{E2M1} & \phantom{0}45 & 41.47 & \multicolumn{1}{c|}{\cellcolor{DarkGreen!25}{\textcolor{DarkGreen}{\textbf{62.07}}}} & \code{E2M1} & \code{E2M1} & \phantom{0}43 \\
\code{3$\times$6} & \phantom{0}44 & \cellcolor{DarkGreen!25}{\textcolor{DarkGreen}{\textbf{67.08}}} & \multicolumn{1}{c|}{66.14} & \code{E1M1} & \code{E2M3} & \phantom{0}57 & \cellcolor{DarkGreen!25}{\textcolor{DarkGreen}{\textbf{48.38}}} & \multicolumn{1}{c|}{41.28} & \code{E1M1} & \code{E2M3} & \phantom{0}57 & 71.98 & \multicolumn{1}{c|}{\cellcolor{DarkGreen!25}{\textcolor{DarkGreen}{\textbf{72.06}}}} & \code{E1M1} & \code{E3M2} & \phantom{0}75 \\
\code{4$\times$5} & \phantom{0}50 & 67.67 & \multicolumn{1}{c|}{\cellcolor{DarkGreen!25}{\textcolor{DarkGreen}{\textbf{68.60}}}} & \code{E2M1} & \code{E2M2} & \phantom{0}57 & 50.13 & \multicolumn{1}{c|}{\cellcolor{DarkGreen!25}{\textcolor{DarkGreen}{\textbf{54.30}}}} & \code{E1M2} & \code{E2M2} & \phantom{0}58 & 68.62 & \multicolumn{1}{c|}{\cellcolor{DarkGreen!25}{\textcolor{DarkGreen}{\textbf{74.01}}}} & \code{E2M1} & \code{E3M1} & \phantom{0}78 \\
\code{3$\times$7} & \phantom{0}51 & \cellcolor{DarkGreen!25}{\textcolor{DarkGreen}{\textbf{67.49}}} & \multicolumn{1}{c|}{66.30} & \code{E1M1} & \code{E3M3} & \phantom{0}74 & \cellcolor{DarkGreen!25}{\textcolor{DarkGreen}{\textbf{56.59}}} & \multicolumn{1}{c|}{45.63} & \code{E1M1} & \code{E2M4} & \phantom{0}60 & \cellcolor{DarkGreen!25}{\textcolor{DarkGreen}{\textbf{73.45}}} & \multicolumn{1}{c|}{72.46} & \code{E1M1} & \code{E3M3} & \phantom{0}74 \\
\code{3$\times$8} & \phantom{0}57 & \cellcolor{DarkGreen!25}{\textcolor{DarkGreen}{\textbf{67.69}}} & \multicolumn{1}{c|}{66.51} & \code{E1M1} & \code{E4M3} & 107 & \cellcolor{DarkGreen!25}{\textcolor{DarkGreen}{\textbf{59.45}}} & \multicolumn{1}{c|}{47.63} & \code{E1M1} & \code{E3M4} & \phantom{0}86 & \cellcolor{DarkGreen!25}{\textcolor{DarkGreen}{\textbf{74.04}}} & \multicolumn{1}{c|}{72.56} & \code{E1M1} & \code{E3M4} & \phantom{0}86 \\
\code{4$\times$6} & \phantom{0}56 & 68.79 & \multicolumn{1}{c|}{\cellcolor{DarkGreen!25}{\textcolor{DarkGreen}{\textbf{69.13}}}} & \code{E2M1} & \code{E2M3} & \phantom{0}64 & {62.44} & \multicolumn{1}{c|}{\cellcolor{DarkGreen!25}{\textcolor{DarkGreen}{\textbf{64.09}}}} & \code{E2M1} & \code{E2M3} & \phantom{0}64 & 74.08 & \multicolumn{1}{c|}{\cellcolor{DarkGreen!25}{\textcolor{DarkGreen}{\textbf{75.27}}}} & \code{E2M1} & \code{E3M2} & \phantom{0}83 \\
\code{5$\times$5} & \phantom{0}55 & 68.02 & \multicolumn{1}{c|}{\cellcolor{DarkGreen!25}{\textcolor{DarkGreen}{\textbf{69.02}}}} & \code{E2M2} & \code{E2M2} & \phantom{0}65 & 56.42 & \multicolumn{1}{c|}{\cellcolor{DarkGreen!25}{\textcolor{DarkGreen}{\textbf{60.30}}}} & \code{E2M2} & \code{E2M2} & \phantom{0}65 & 69.44 & \multicolumn{1}{c|}{\cellcolor{DarkGreen!25}{\textcolor{DarkGreen}{\textbf{74.64}}}} & \code{E2M2} & \code{E3M1} & \phantom{0}83 \\
\code{4$\times$7} & \phantom{0}64 & 69.05 & \multicolumn{1}{c|}{\cellcolor{DarkGreen!25}{\textcolor{DarkGreen}{\textbf{69.25}}}} & \code{E2M1} & \code{E2M4} & \phantom{0}78 & 66.04 & \multicolumn{1}{c|}{\cellcolor{DarkGreen!25}{\textcolor{DarkGreen}{\textbf{66.81}}}} & \code{E2M1} & \code{E2M4} & \phantom{0}78 & 74.91 & \multicolumn{1}{c|}{\cellcolor{DarkGreen!25}{\textcolor{DarkGreen}{\textbf{75.49}}}} & \code{E2M1} & \code{E3M3} & 111 \\
\code{5$\times$6} & \phantom{0}63 & 69.20 & \multicolumn{1}{c|}{\cellcolor{DarkGreen!25}{\textcolor{DarkGreen}{\textbf{69.47}}}} & \code{E1M3} & \code{E2M3} & \phantom{0}82 & 66.64 & \multicolumn{1}{c|}{\cellcolor{DarkGreen!25}{\textcolor{DarkGreen}{\textbf{67.96}}}} & \code{E2M2} & \code{E2M3} & \phantom{0}89 & 74.39 & \multicolumn{1}{c|}{\cellcolor{DarkGreen!25}{\textcolor{DarkGreen}{\textbf{75.54}}}} & \code{E2M2} & \code{E3M2} & \phantom{0}95 \\
\code{4$\times$8} & \phantom{0}67 & 69.24 & \multicolumn{1}{c|}{\cellcolor{DarkGreen!25}{\textcolor{DarkGreen}{\textbf{69.37}}}} & \code{E2M1} & \code{E2M5} & \phantom{0}78 & \cellcolor{DarkGreen!25}{\textcolor{DarkGreen}{\textbf{68.05}}} & \multicolumn{1}{c|}{67.93} & \code{E2M1} & \code{E2M5} & \phantom{0}78 & 75.45 & \multicolumn{1}{c|}{\cellcolor{DarkGreen!25}{\textcolor{DarkGreen}{\textbf{75.58}}}} & \code{E2M1} & \code{E3M4} & 110 \\
\code{5$\times$7} & \phantom{0}71 & 69.47 & \multicolumn{1}{c|}{\cellcolor{DarkGreen!25}{\textcolor{DarkGreen}{\textbf{69.56}}}} & \code{E1M3} & \code{E2M4} & 103 & 68.91 & \multicolumn{1}{c|}{\cellcolor{DarkGreen!25}{\textcolor{DarkGreen}{\textbf{70.16}}}} & \code{E2M2} & \code{E2M4} & \phantom{0}87 & 75.29 & \multicolumn{1}{c|}{\cellcolor{DarkGreen!25}{\textcolor{DarkGreen}{\textbf{75.75}}}} & \code{E2M2} & \code{E3M3} & 118 \\
\code{6$\times$6} & \phantom{0}72 & 69.18 & \multicolumn{1}{c|}{\cellcolor{DarkGreen!25}{\textcolor{DarkGreen}{\textbf{69.55}}}} & \code{E2M3} & \code{E2M3} & \phantom{0}89 & 67.80 & \multicolumn{1}{c|}{\cellcolor{DarkGreen!25}{\textcolor{DarkGreen}{\textbf{68.30}}}} & \code{E2M3} & \code{E2M3} & \phantom{0}89 & 74.55 & \multicolumn{1}{c|}{\cellcolor{DarkGreen!25}{\textcolor{DarkGreen}{\textbf{75.64}}}} & \code{E3M2} & \code{E3M2} & 110 \\
\code{5$\times$8} & \phantom{0}78 & \cellcolor{DarkGreen!25}{\textcolor{DarkGreen}{\textbf{69.66}}} & \multicolumn{1}{c|}{69.60} & \code{E1M3} & \code{E2M5} & \phantom{0}86 & 70.26 & \multicolumn{1}{c|}{\cellcolor{DarkGreen!25}{\textcolor{DarkGreen}{\textbf{70.63}}}} & \code{E2M2} & \code{E2M5} & 101 & 75.74 & \multicolumn{1}{c|}{\cellcolor{DarkGreen!25}{\textcolor{DarkGreen}{\textbf{75.83}}}} & \code{E2M2} & \code{E3M4} & 154 \\
\code{6$\times$7} & \phantom{0}78 & 69.54 & \multicolumn{1}{c|}{\cellcolor{DarkGreen!25}{\textcolor{DarkGreen}{\textbf{69.60}}}} & \code{E2M3} & \code{E3M3} & 122 & 69.98 & \multicolumn{1}{c|}{\cellcolor{DarkGreen!25}{\textcolor{DarkGreen}{\textbf{70.62}}}} & \code{E2M3} & \code{E2M4} & 119 & 75.26 & \multicolumn{1}{c|}{\cellcolor{DarkGreen!25}{\textcolor{DarkGreen}{\textbf{75.84}}}} & \code{E2M3} & \code{E3M3} & 122 \\
\code{6$\times$8} & \phantom{0}87 & \cellcolor{AppleGreen!25}{69.63} & \multicolumn{1}{c|}{\cellcolor{DarkGreen!25}{\textcolor{DarkGreen}{\textbf{69.67}}}} & \code{E3M2} & \code{E3M4} & 134 & 71.06 & \multicolumn{1}{c|}{\cellcolor{DarkGreen!25}{\textcolor{DarkGreen}{\textbf{71.24}}}} & \code{E2M3} & \code{E2M5} & 102 & 75.70 & \multicolumn{1}{c|}{\cellcolor{DarkGreen!25}{\textcolor{DarkGreen}{\textbf{75.89}}}} & \code{E1M4} & \code{E3M4} & 110 \\
\code{7$\times$7} & \phantom{0}92 & \cellcolor{DarkGreen!25}{\textcolor{DarkGreen}{\textbf{69.65}}} & \multicolumn{1}{c|}{\cellcolor{AppleGreen!25}{69.63}} & \code{E4M2} & \code{E2M4} & 174 & 70.31 & \multicolumn{1}{c|}{\cellcolor{DarkGreen!25}{\textcolor{DarkGreen}{\textbf{70.77}}}} & \code{E1M5} & \code{E2M4} & 107 & 75.30 & \multicolumn{1}{c|}{\cellcolor{DarkGreen!25}{\textcolor{DarkGreen}{\textbf{75.79}}}} & \code{E2M4} & \code{E3M3} & 131 \\
\code{7$\times$8} & 110 & \cellcolor{AppleGreen!25}{69.65} & \multicolumn{1}{c|}{\cellcolor{DarkGreen!25}{\textcolor{DarkGreen}{\textbf{69.69}}}} & \code{E2M4} & \code{E3M4} & 130 & 71.30 & \multicolumn{1}{c|}{\cellcolor{DarkGreen!25}{\textcolor{DarkGreen}{\textbf{71.38}}}} & \code{E2M4} & \code{E2M5} & 121 & 75.72 & \multicolumn{1}{c|}{\cellcolor{DarkGreen!25}{\textcolor{DarkGreen}{\textbf{75.90}}}} & \code{E4M2} & \code{E3M4} & 183 \\
\code{8$\times$8} & 116 & \cellcolor{DarkGreen!25}{\textcolor{DarkGreen}{\textbf{69.71}}} & \multicolumn{1}{c|}{\cellcolor{AppleGreen!25}{69.68}} & \code{E3M4} & \code{E2M5} & 147 & 71.36 & \multicolumn{1}{c|}{\cellcolor{DarkGreen!25}{\textcolor{DarkGreen}{\textbf{71.52}}}} & \code{E2M5} & \code{E2M5} & 133 & 75.79 & \multicolumn{1}{c|}{\cellcolor{DarkGreen!25}{\textcolor{DarkGreen}{\textbf{75.89}}}} & \code{E4M3} & \code{E4M3} & 191 \\ \bottomrule
\end{tabular}
\vskip -0.2in
\end{table*}

Both MACs feature parameterizable operand formats, which we adapt to align with our experiments. Additionally, we set their pipeline depth to a fixed value of two, 
ensuring reasonable operating frequencies. Notably, we omit considerations for the hardware costs associated with converting the accumulated values in the minifloat MAC back into the floating-point format, as this operation can typically be merged into the subsequent activation function as by implementations based on thresholding~\cite{blott2018finn}. Note that our model accuracy assessments are conducted with accumulators in \code{FP32}. Given the limited precision of the minifloat formats under consideration, this choice adequately ensures the validity of our results.


As the integer MAC accumulates products in a straightforward manner, its accumulator width is $\text{r}_a+\text{r}_b+\lceil\log_2n\rceil+1$, where $\text{r}_a$ and $\text{r}_b$ are the operand widths, and $n$ is the maximum number of supported products~\cite{a2q}. However, as the minifloat MAC necessitates shifting the significand product before accumulation, its accumulator width grows exponentially with the exponent bit-widths of the operand formats and is computed as $2^{e_a}+m_a+2^{e_b}+m_b+\lceil\log_2n\rceil-1$, where $e_a$ and $e_b$ are the exponent bit-widths and $m_a$ and $m_b$ are the mantissa bit-widths~\cite{de2017design}. This leads us to anticipate that minifloat MACs for operands with wide exponents may be uneconomical. For our experiments, we choose $n$ as the maximum dot product size 
found in ResNet-18\footnote{MobileNetV2 and ViT-B-32 feature maximum dot product sizes of $1280$ and $3072$, meaning they require one or two fewer bits in their accumulators. These differences have negligible impact on the LUT utilization of our MACs.}, $4608$.


\section{Experimental Setup} \label{sec:experiments}

We focus our experiments on three deep learning-based vision models: ResNet-18~\cite{7780459}, MobileNetV2~\cite{8578572}, and ViT-B-32~\cite{dosovitskiy2020vit} on the ImageNet-1K dataset~\cite{deng2009imagenet}. Our code is implemented in PyTorch, utilizing the Brevitas library for post-training integer and minifloat quantization~\cite{brevitas}. We restrict our experiments to signed representations with $z$ set to $0$, a choice widely adopted to mitigate computational overhead during the inference process~\cite{tuq}. We quantize the first and last layers in the \code{INT8} and \code{E3M4} formats for integer and minifloat quantization, respectively, since these layers are sensitive to quantization~\cite{Zhao2019ImprovingNN, Choi2018PACTPC}. In line with previous work~\cite{pmlr-v139-hubara21a}, we use a small calibration dataset of 1000 images for activation calibration, bias correction, learned rounding, and GPTQ.

We consider data types spanning 3 to 8-bit formats for both weights and activations in convolutional, linear, and general matrix multiplication layers. 
We keep the same number of mantissa and exponent bits across all network layers. We conduct a thorough \emph{design-space exploration}, experimenting with all combinations of mantissa and exponent bits for a given target bit-width, and report the configurations corresponding to the best model accuracy. Our experiments encompass both per-tensor and per-channel weight scaling with scaling factors preserved in \code{FP32} format. For a given experiment, we apply either GPTQ or learned rounding but not both, while all other PTQ methods can be enabled concurrently. 

We investigate the accuracy-hardware trade-offs of our quantized models using two hardware cost metrics.
The first metric is the \textbf{memory footprint}, defined with respect to the weight bit-width in a quantized model to compare the relative costs of different precision configurations. 
The second metric is \textbf{look-up table (LUT) utilization} of a MAC on the FPGA for a particular set of weight and activation formats and dot product size. 
Prior research efforts employ 
similar metrics for memory and arithmetic density~\cite{NEURIPS2020_747e32ab, a2q}.
We choose to focus solely on LUTs rather than hardened digital signal processing slices (DSPs) for two reasons: firstly, DSPs are relatively rare in Versal\filterblind{™} FPGAs (roughly, one per 200-400 LUTs~\cite{xilinx2023ds950}) and their lack of native support for minifloat formats would give integer designs an unintended upper hand; and secondly, using one metric simplifies comparisons. All included results are post-implementation utilization numbers produced with default settings in Vivado\filterblind{™} 2023.1.

\section{Evaluation \& Discussion}

\subsection{Impact on Model Accuracy}
We first discuss the Top-1 model accuracy (\%) for various precision configurations and 
present the best-case results across all post-training optimization techniques in Table~\ref{fp_int_acc}.
We observe that integer quantization outperforms minifloat quantization with 3-bit weights and 3- to 4-bit activations across all three models. However, this performance gap diminishes as the weight precision increases to 4 bits and beyond. For configurations where weights and activations are set to 4 and 5~bits, respectively, and above, minifloat quantization outperforms integer quantization by up to $6\%$ for models such as ViT-B-32. 
Notably, with 4-bit weights and 8-bit activations, both integer and minifloat representations closely approach the accuracy of the full-precision model. Upon individual model analysis, {we conclude that while traditional convolutional models like ResNet-18 perform equally well with integer and minifloat quantization, more complex models such as MobileNetV2 and ViT-B-32, characterized by the presence of outliers in their weight and activation distributions, benefit from minifloat quantization due to its larger dynamic range.}


\begin{figure*}[t]
\centering
    \begin{minipage}[t]{0.32\linewidth}
        \includegraphics[width=\linewidth]{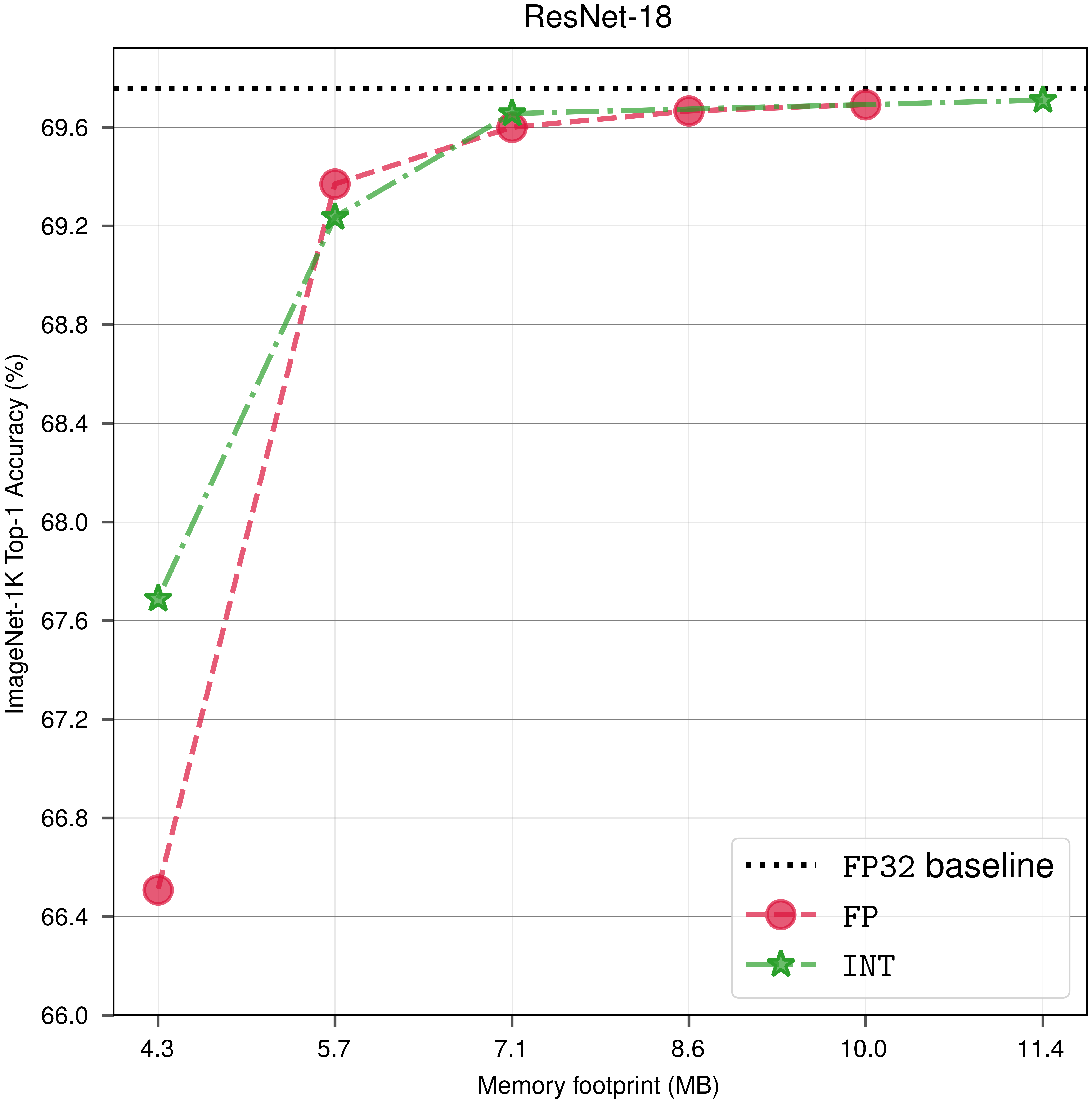}
    \end{minipage}%
        \hfill%
    \begin{minipage}[t]{0.32\linewidth}
        \includegraphics[width=\linewidth]{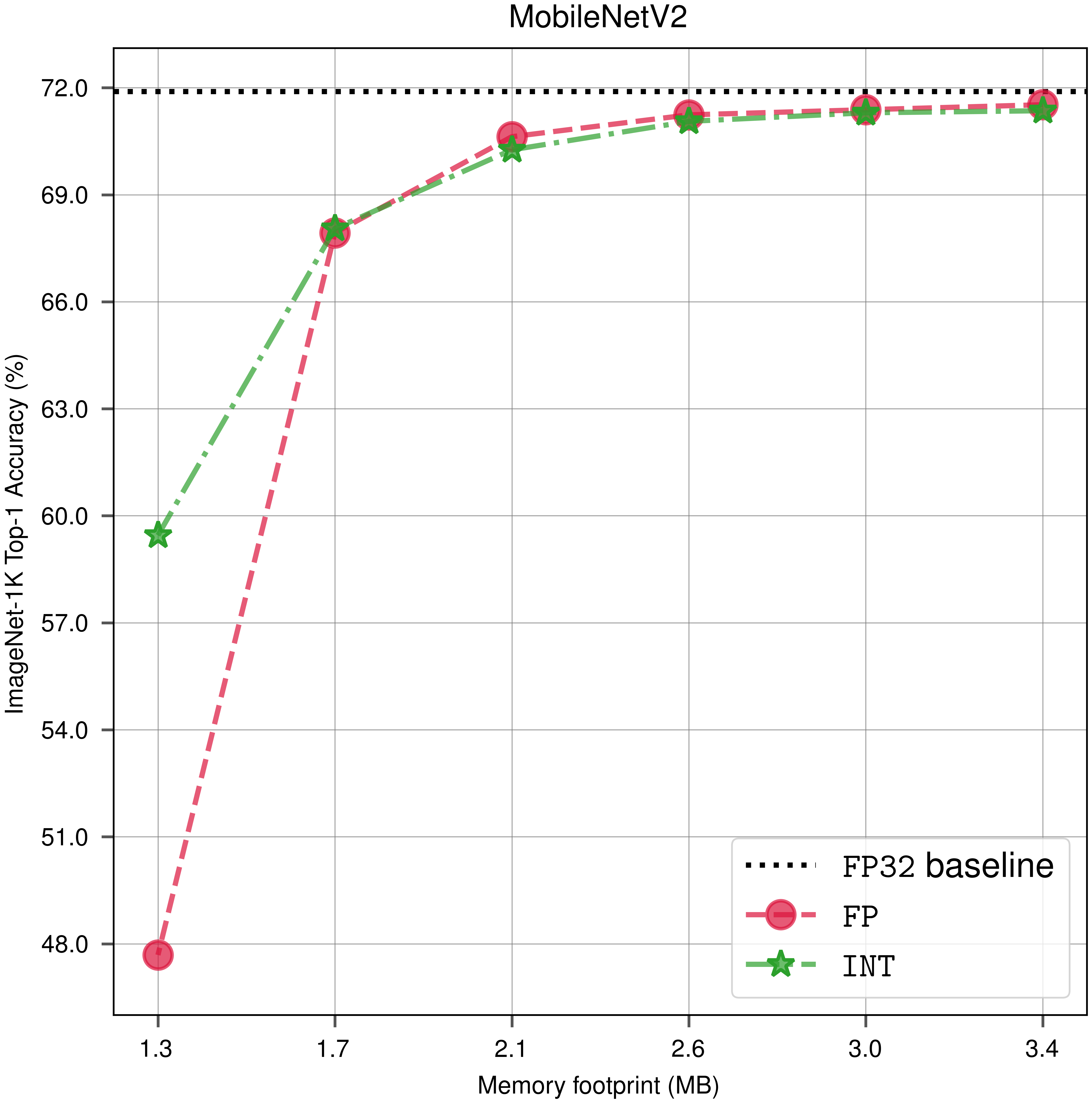}
    \end{minipage} 
    \hfill%
    \begin{minipage}[t]{0.32\linewidth}
        \includegraphics[width=\linewidth]{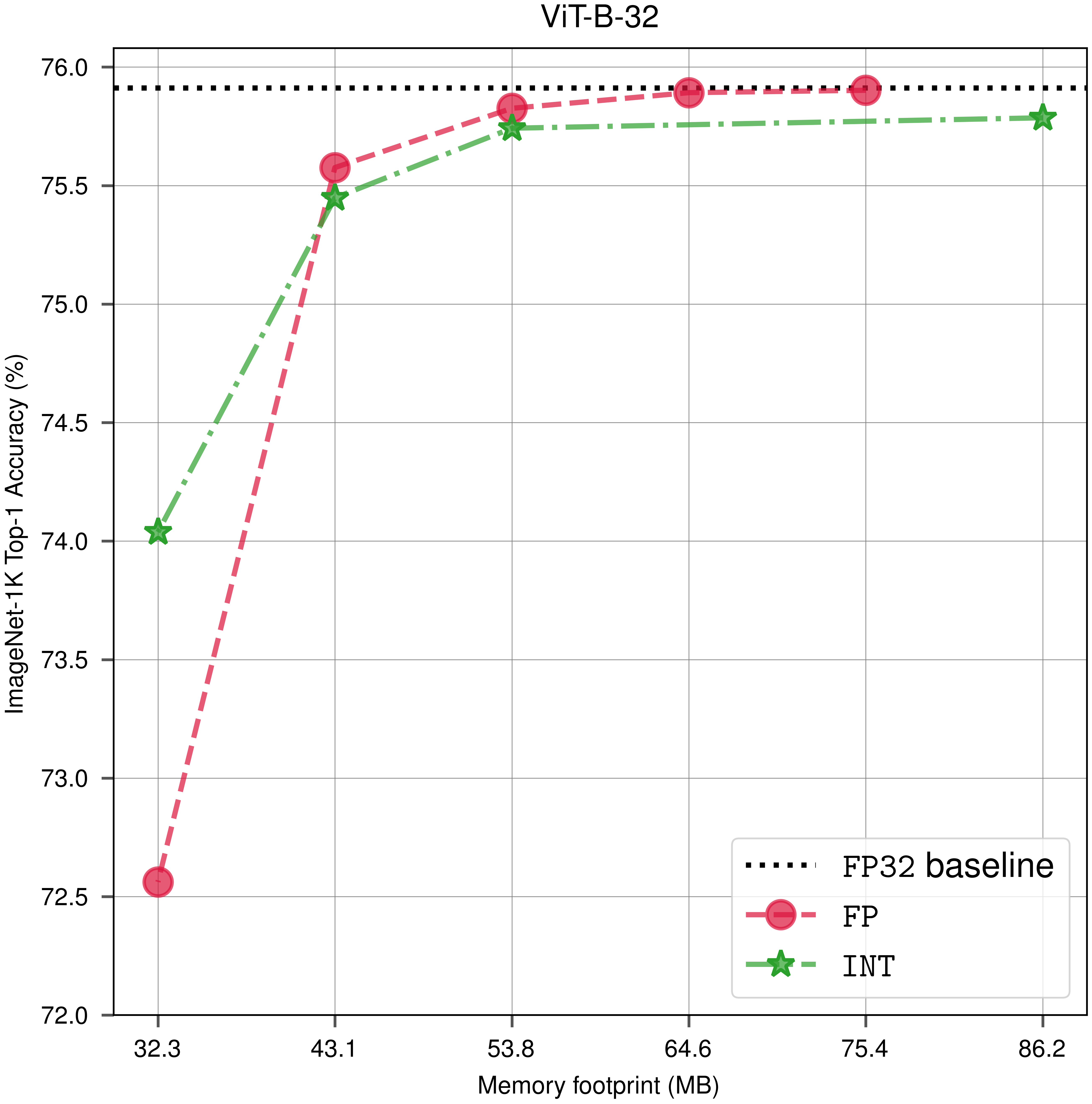}
    \end{minipage}
    \caption{Trade-off analysis between \underline{model accuracy} (\%) and \underline{memory footprint} for {integers} (\code{INT}) and {minifloats} (\code{FP}). 
    }
    \label{fig:overall_bitwidth}
           \vskip -0.15in
\end{figure*}

\begin{figure*}[t]
    \begin{minipage}[t]{0.325\linewidth}
        \includegraphics[width=\linewidth]{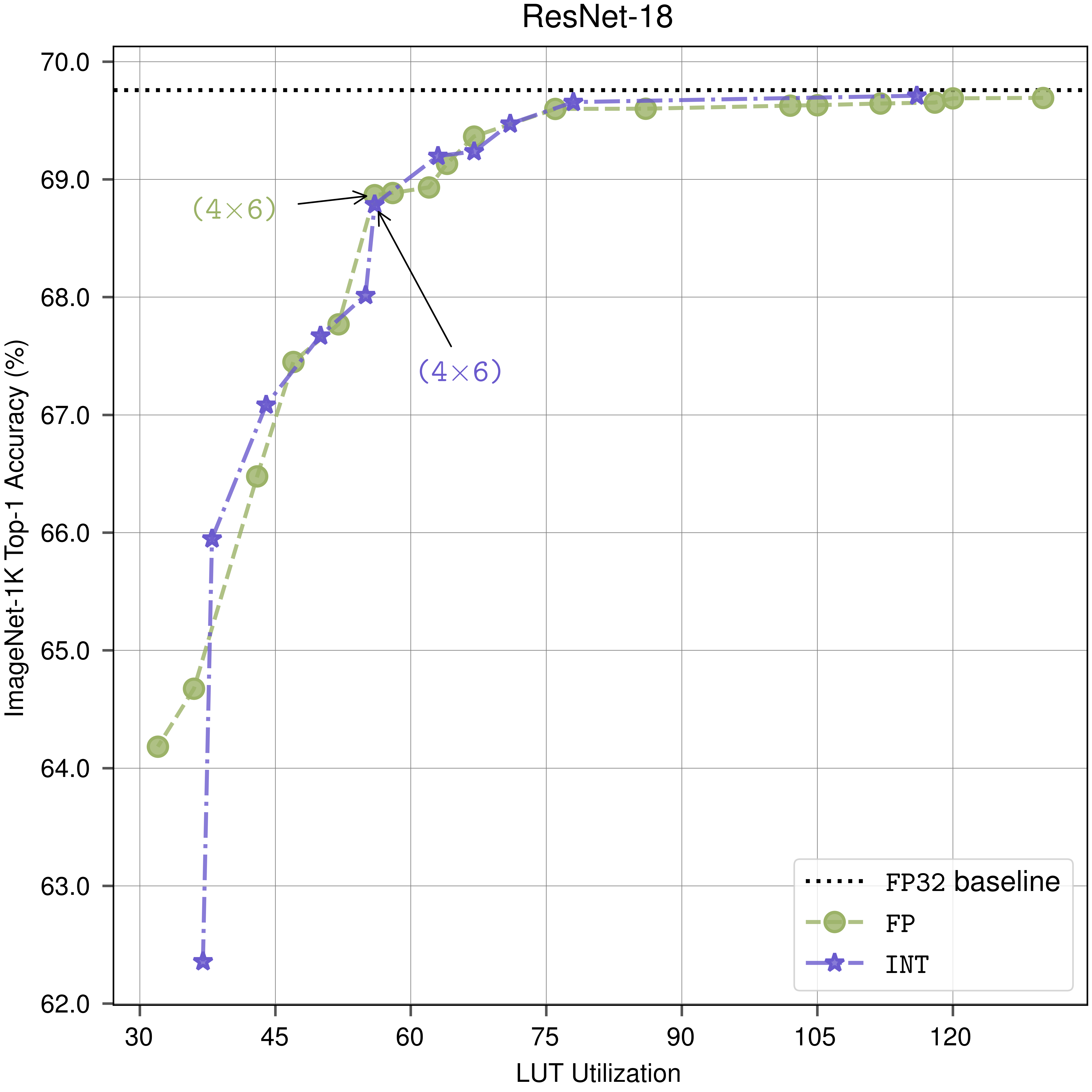}
    \end{minipage}%
        \hfill%
    \begin{minipage}[t]{0.325\linewidth}
        \includegraphics[width=\linewidth]{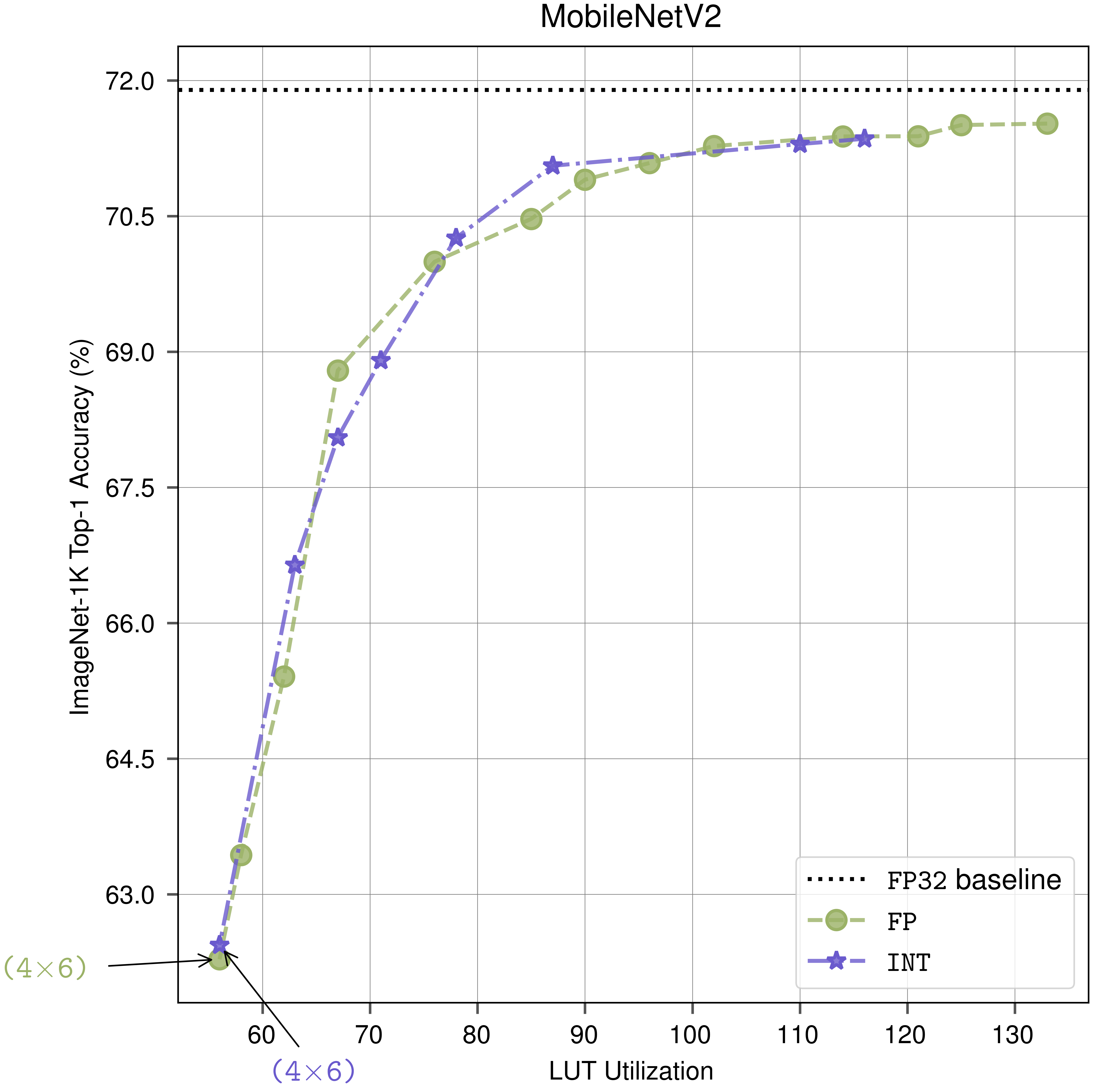}
    \end{minipage} 
    \hfill%
    \begin{minipage}[t]{0.325\linewidth}
        \includegraphics[width=\linewidth]{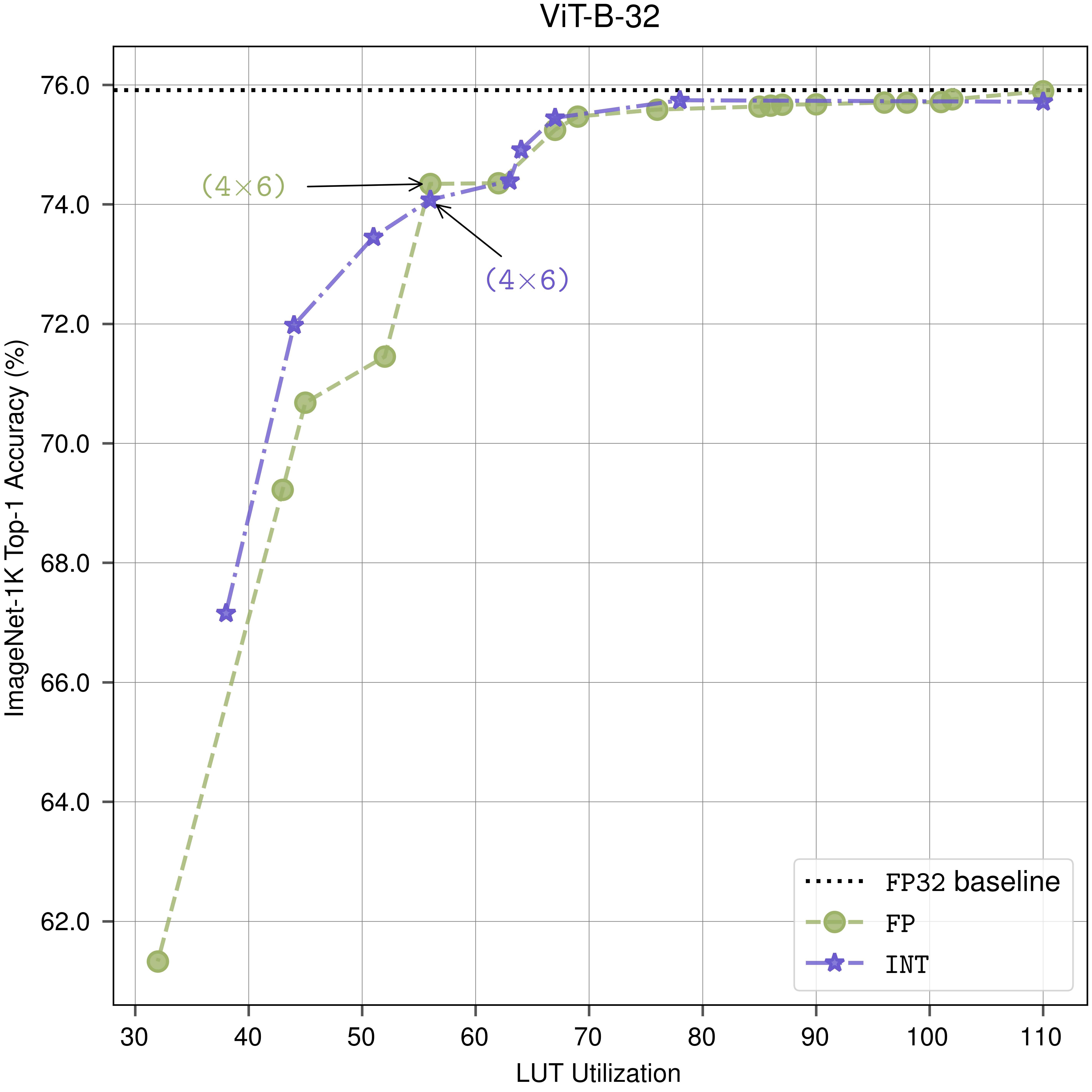}
    \end{minipage}
    \caption{Trade-off analysis between \underline{model accuracy} (\%) and \underline{\#LUT utilization} for {integers} (\code{INT}) and {minifloats} (\code{FP}). Points where the gap between the two formats converge are labeled with their corresponding bit-width configurations as (\code{W}$\times$\code{A}).}
    \label{fig:overall_LUTs}
           \vskip -0.2in
\end{figure*}

\subsection{Optimal Minifloat Formats}
Table~\ref{fp_int_acc} showcases the minifloat formats that achieve the best model accuracy at each weight and activation bit-width configuration. Given that these exponent and mantissa bit-width choices directly influence hardware resource utilization costs, a closer examination of their values is essential. As anticipated, when considering weights, most configurations comprise at least two exponent bits.
Conversely, a consistent trend emerges for activations, with most models requiring exponents of two or more bits. We note a substantial disparity in the necessary representations of the weights and activations across all three models. Such design requirements can only be accommodated on re-programmable fabric such as FPGAs.

\subsection{Impact on Memory Footprint}
We illustrate the trade-off between model accuracy and memory footprint for 3- to 8-bit formats through Pareto curves in Fig.~\ref{fig:overall_bitwidth}. The black dotted lines represent the accuracy of the three models in \code{FP32}. 
Both integer and minifloat formats exhibit sub-optimal performance at 3~bits, particularly for MobileNetV2 and ViT-B-32, where 3-bit integers marginally outperform minifloats. For ResNet-18, both formats exhibit similar performance at different weight bit-widths. For complex networks such as MobileNetV2 and ViT-B-32, minifloats dominate integers at higher bit-widths since they can easily accommodate outliers. For instance, for ViT-B-32, minifloat quantization can attain a remarkable accuracy of $75.89\%$ with an impressive $8\times$ compression rate compared to the full-precision model (calculated as $32/4$).


\subsection{Impact on FPGA Resource Utilization}
Unlike our previous analysis, which focused on determining the optimal model accuracy at varying memory budgets, this section shifts focus to identifying the best model accuracy with respect to the number of LUTs. 
For the integer MAC, a lower-precision configuration generally utilizes fewer LUTs. However, the resource utilization of a minifloat MAC depends on its distribution of exponent and mantissa bits, especially due to its shifter and, possibly, long accumulator. Observant readers will notice that, in some cases, increasing the bit-width of minifloat formats can lead to reduced LUT utilization. For example, Table~\ref{fp_int_acc} shows that the \code{3}$\times$\code{7} configuration outperforms \code{3}$\times$\code{6} for ViT-B-32. We attribute this to some formats fitting the underlying fabric better, e.g., the seven-bit \code{E3M3} format~\cite{Xilinx}, and to some degree of randomness in the optimizations performed by the synthesis tool.

From the curves in Fig.~\ref{fig:overall_LUTs}, we observe that integer quantization tends to outperform minifloat quantization when the LUT utilization is very limited, albeit with significant accuracy degradation as compared to the \code{FP32} baseline. However, as the resource budget increases, both integer and minifloat accuracies show an upward trend and eventually converge, as indicated by the labeled bit-width configurations.
In the case of ResNet-18, we notice an accuracy difference of $<$$1\%$ when the resource budget is around 45~LUTs. This gap drops significantly as the resource budget surpasses 50~LUTs. Similarly, for MobileNetV2, minifloat quantization performs on par with integer representations, where the margin is less than $0.1\%$ for resource budgets exceeding 64~LUTs. Results for ViT-B-32 show a trend similar to ResNet-18, with the gap narrowing as the number of LUTs exceeds 60. 


\section{Conclusion}
In this study, we present a minifloat quantization approach for bit-widths below 8~bits. We conduct a thorough examination of the impact of various PTQ techniques on minifloats and integers with reduced precision spanning from 3 to 8~bits, for both weights and activations. Our research centers on assessing trade-offs between accuracy and hardware resource utilization for ResNet-18, MobileNetV2, and ViT-B-32 models and two bespoke MAC designs implemented on an FPGA. Our experiments demonstrate the relevance of low-precision minifloat quantization, especially in terms of memory footprint at 4~bits or above, while conceding that integer quantization still dominates along the MAC resource utilization Pareto frontier. Our experiments primarily focus on vision models and assume a uniform choice of data formats for all weights and activations in a given model, which limits the design space of configurations that can be explored. We plan to address these limitations in our future work.

\section {Acknowledgments}
We thank the anonymous reviewers, Ian Colbert (AMD), and Dan Wu (NUS) for their valuable feedback. Shivam Aggarwal's work is partially supported by the
National Research Foundation, Singapore, under its Competitive Research Programme Award NRF-CRP23-2019-0003 and Singapore Ministry of Education Academic Research Fund T1 251RES1905. Hans Jakob Damsgaard acknowledges funding from European Union's Horizon 2020 Research and Innovation Programme under the Marie Sk\l{}odowska Curie grant agreement No. $956090$ (APROPOS: Approximate Computing for Power and Energy Optimisation, http://www.apropos-itn.eu/). 

\bibliographystyle{IEEEtran}
\bibliography{bib}

\end{document}